\documentclass[letterpaper]{article} 
\usepackage{aaai23}  
\usepackage{amsmath}
\usepackage{amsfonts}
\usepackage{amssymb}
\usepackage{times}  
\usepackage{helvet}  
\usepackage{courier}  
\usepackage[hyphens]{url}  
\usepackage{graphicx} 
\usepackage{natbib}  
\usepackage{caption} 
\usepackage{algorithm}
\usepackage{algorithmic}

\usepackage{newfloat}
\usepackage{listings}
\DeclareCaptionStyle{ruled}{labelfont=normalfont,labelsep=colon,strut=off} 
\floatstyle{ruled}
\newfloat{listing}{tb}{lst}{}
\floatname{listing}{Listing}
\title{A Multimodal Approach for Advanced Pest Detection and Classification}

\author{
    Jinli Duan\textsuperscript{\rm 1}\thanks{These authors contributed equally to this work.}, 
    Haoyu Ding\textsuperscript{\rm 1}\footnotemark[1], 
    Sung Kim\textsuperscript{\rm 1}\footnotemark[1]
}
\affiliations {
    \textsuperscript{\rm 1} NYU Tandon\\
    jd5374@nyu.edu, haoyu\_ding@outlook.com, sk9623@nyu.edu
}

\begin{document}

\maketitle

\begin{abstract}
This paper presents a novel multi modal deep learning framework for enhanced agricultural pest detection, combining tiny-BERT's natural language processing with R-CNN and ResNet-18's image processing. Addressing limitations of traditional CNN-based visual methods, this approach integrates textual context for more accurate pest identification. The R-CNN and ResNet-18 integration tackles deep CNN issues like vanishing gradients, while tiny-BERT ensures computational efficiency. Employing ensemble learning with linear regression and random forest models, the framework demonstrates superior discriminate ability, as shown in ROC and AUC analyses. This multi modal approach, blending text and image data, significantly boosts pest detection in agriculture. The study highlights the potential of multi modal deep learning in complex real-world scenarios, suggesting future expansions in diversity of datasets, advanced data augmentation, and cross-modal attention mechanisms to enhance model performance.

\end{abstract}
\section{Introduction}

Pest infestations in agricultural crops are a concern to the world with some far reaching consequences, and it can cause a lower quality of agricultural product or permanent damage and compromise lead to reduction significantly. Which has implications for food security, economic stability. In the rapidly evolving agricultural technology and deep learning techniques could migrate this problem. In order to prevent or migrate the growth of the agricultural pests, detecting pests precisely and timelessly is crucial.

Traditional image detection methods which are also used for the pest detection, particularly relying on single methods such as Convolutions Neural Network (CNN) models to primarily focus on visual data analysis \cite{alzubaidi2021review}. Using R-CNN models have shown efficacy in image recognition tasks Image detection\cite{bharati2020deep}; however this reliance on solely visual limits there could be possible improvement through integrating a model from others.

This report introduces a multi-modal deep learning approach, combining the strength of tiny-BERT which and R-CNN with ResNet-18 as backbone, which has exceptional performance in natural language processing. The combination of visual data processing and natural language processing can increase the possible positive differences \cite{jiao2019tinybert}. 

The R-CNN with ResNet-18(Residual Network-18) is used instead of traditional CNN architectures due to innovative use of residual network learning \cite{targ2016resnet}. This effectively addresses the vanishing gradient problems, and over fitting issues can be occurred in deep CNN. thereby allowing the construction of a much deeper, yet more accurate and efficient model for this project. Tiny-Bert is employed over the standard Transformer and regular BERT models due to its optimized architecture. It allows while maintaining a performance analogous to its predecessors in natural language processing tasks, markedly reduces computational and consumption of the resources. Therefore, it can reduce the size significantly while maintaining much of the performance by giving slight trade off in accuracy.

Notwithstanding the potential for yielding favorable outcomes, there remains consideration of its applicability and effectiveness. The process of collecting image-labeled data presents a certain degree of tedious; however it is relatively more straightforward contrasted with  language input data sets. The data-sets for fine tuning  tiny-BERT dominantly derived from LLAVA which necessitated the use of a high performance graphics processing unit and considerable investment in time for data-sets \cite{liu2023improved}. Despite these challenges, the result of both combinations of tiny-BERT and RCNN augmented by ResNet-18 as backbone, yielded satisfying outcomes, and demonstrably distinctions of integrative approach.

\section{Related Work}

The domain of image detection has significant advancement with CNNs. Prominent among these developments are models as ResNet, VGG, GoogleNet, Alexnet and other CNN methods are used as backbone. According to Nyarko The models AlexNet, ResNet-50, and Inception-V3 were employed, with ResNet-50 achieving the highest accuracy in facial recognition via mouth detection, recording 100 percent accuracy in training and 97.5 percent in testing. This performance notably surpassed that of the other models \cite{nyarko2022comparative}. The ResNet-50 demonstrated notably superior performance compared to the other models; therefore, this model is suited for the backbone of the R-CNN image classification model. Even Though there is a slight trade off for ResNet-18 because of the efficiency this model suits for the method. Zhang Claimed that CNN models with large data-sets and retraining with smaller data-set is effective to solve large and diverse data sets in the pest recognition field \cite{zhang2023multimodal}. the MSR-RCNN method demonstrates significant improvements in pest detection by addressing the specific challenges of pest object characteristics, such as small size, varied scales, and high similarity \cite{teng2022msr}. Employing the R-CNN framework demonstrates considerable efficacy in the identification and localization of pests within image data sets.

The transformer model revolutionized Natural Language Processing due to its unique structure and efficiency because the attention mechanism word of sequence does not matter to process step by step \cite{vaswani2017attention}. And this parallel processing significantly speeds up training to improve.
\cite{redmon2016you}. It is used as an actual object detection model with a CNN. unlike these 2 stage model.  In our proposed model, we integrate Tiny-BERT as the linguistic processing component to analyze textual data extracted from images. This integration is designed to synergize with our R-CNN framework, which is fortified with a ResNet-18 backbone, thereby enhancing the overall accuracy of the system.

multi modal enhancing their analytical robustness and accuracy in complex tasks. 
introduce MMCNN-MIML, a deep multi-modal CNN for MIML image classification, which outperforms on benchmarks due to its unique approach: it groups related class labels for better feature learning, represents images as bags of visual instances rather than single entities, and uses group descriptions to enhance discrimination of visually similar objects in different groups \cite{song2018deep}. However, rather than directly integrating the Multi-Modal approach within the CNN model, our methodology entails independently scoring the Tiny-BERT model and subsequently amalgamating these scores in the final stage to achieve enhanced accuracy.

\section{Related Methods}

In this section, we discuss several important metrics, loss functions, regularization techniques, and model fusion methods commonly used in our project.

\subsection{Metrics}

\begin{enumerate}
    \item \textbf{F1 Score}:
    The F1 Score is a widely used metric for evaluating classification models. In our case, we used F1 scores in CNN-Resnet scores by balancing the precision and recall to provides a more comprehensive performance assessment. The F1 score combines precision and recall and is defined as:
    \begin{equation}
    F1 = 2 \cdot \frac{\text{Precision} \cdot \text{Recall}}{\text{Precision} + \text{Recall}}
    \end{equation}

    \item \textbf{Accuracy}:
    Accuracy measures the ratio of correct predictions to the total number of predictions. This has been used as Metrics for both CNN-Resnet model and Fine-tuned Tiny-BERT and is defined as:
    \begin{equation}
    \text{Accuracy} = \frac{\text{Number of Correct Predictions}}{\text{Total Number of Predictions}}
    \end{equation}

    \item \textbf{AUC (Area Under the ROC Curve)}:
    The Area Under the Curve (AUC) metric measures a model's ranking ability, making it particularly suitable for ranking-oriented tasks. AUC is insensitive to the balance of positive and negative samples, allowing for reasonable evaluation even in imbalanced scenarios. In contrast, other metrics like precision, recall, and F1 score can vary based on the threshold set for distinguishing between positive and negative samples. Since AUC doesn't require a manual threshold setting, it offers a holistic measurement approach. In our dataset, where the number of pest samples is approximately 1/2 higher than that of non-pest samples, AUC proves to be a more appropriate metric. 
\end{enumerate}
\subsection{Loss Functions}

\begin{itemize}
    \item \textbf{Cross-entropy Loss}:
    Cross-entropy loss is frequently employed in classification tasks\cite{CrossEntropy}. For our case, we used Cross entropy loss in the process of both CNN-resnet and Fine-tuning Tiny-Bert for a binary classification purpose. It measures the dissimilarity between the predicted probabilities ($q(x)$) and the true probabilities ($p(x)$) and is defined as:
    \begin{equation}
    H(p,q) = -\sum_{x} p(x) \log q(x)
    \end{equation}
\end{itemize}

\subsection{Regularization Techniques}

\begin{itemize}
    \item \textbf{Dropout}:
    Dropout is a widely adopted regularization technique in neural networks, pioneered by Srivastava et al. \cite{dropout}. This method involves randomly deactivating individual neurons during the training phase with a probability 'p', while keeping them active with a probability of '1-p'. This stochastic process aids in mitigating the risk of overfitting, which is particularly crucial in complex models.
    In the context of our CNN-ResNet architectures, we observed a major issue of overfitting. This was evident from the disparity in performance metrics: the model achieved a remarkable accuracy of 99\% on the training set, yet this accuracy has been  dropped to 82\% on the test set. To address this challenge, we employed the dropout strategy, experimenting with different probabilities. We found that setting the dropout rate to 0.2 yielded the most favorable outcome, striking a balance between learning efficiency and generalization capability. The following table provides a comparative analysis of various dropout rates and their corresponding impacts on the model's performance.
    
    \begin{table}[h]
        \centering
        \begin{tabular}{ccccc}
            \hline
            Methods & without dropout & P=0.1 & P=0.2 & P=0.5 \\
            \hline
            Trainset Accuracy & 99 & 94 & 86 & 54 \\
            Testset Accuracy & 82 & 88 & 90 & 86 \\
            \hline
        \end{tabular}
        \caption{Effects of Dropout on Model Accuracy}
        \label{tab:dropout_effect}
    \end{table}

\end{itemize}

\section{Experiment}

Our approach leverages a hybrid model combining the powerful feature extraction capabilities of CNN-ResNet with the advanced language understanding of a fine-tuned Tiny BERT. 
\subsection{CNN-ResNet Backbone model}

The foundation of our visual recognition part is predicated on the CNN-ResNet architecture, a convolutional neural network renowned for its deep structure and residual connections.

\subsubsection{Model Architecture}
We employ the ResNet50d model, obtained through the \texttt{timm} library, renowned for its pre-trained weights that facilitate the extraction of generic visual features from a vast corpus of image data. To tailor the model to our classification task, the original fully connected layer is replaced with a new linear layer. This layer's output features are equivalent to the number of classes in our dataset, thus adapting the model to the specificities of our task. 

\subsubsection{Data Preprocessing}
Our preprocessing pipeline transforms the input images to a uniform size, enhancing the model's ability to learn scale-invariant features. Further data augmentation is applied to improve model robustness and generalization. This includes random horizontal flips, slight rotations, and automatic contrast adjustments implemented using the \texttt{transforms} and \texttt{albumentations} libraries. 

\subsubsection{Training and Validation}
The model is fine-tuned using custom training and validation functions, which involve computing the loss and accuracy metrics while monitoring performance through the \texttt{MetricMonitor} class. We use cross-entropy loss for training and an AdamW optimizer for weight adjustment. A cosine annealing strategy is applied to the learning rate after each epoch, aiding in the stabilization of the training process over successive iterations.

\subsubsection{Performance Evaluation}
To assess the model's capability in handling multi-class classification, we utilize macro-averaged F1 scores, accuracy and recall rates as our primary performance metrics. In essence, the CNN-ResNet component of our model is integral to capturing the nuanced patterns and textures present in pest image data.
\subsection{Fine-Tuned TinyBERT Model}

In parallel with the visual recognition provided by the CNN-ResNet architecture, our framework incorporates the TinyBERT model for text classification. This choice is motivated by TinyBERT's smaller size and fewer parameters, making it ideal for fine-tuning on our dataset with a limited number of text descriptions generated using LLAVA\cite{llava}.

\subsubsection{Model Architecture}
TinyBERT is a compact version of the larger BERT model\cite{tinybert}, renowned for its effectiveness in natural language processing tasks. We utilize the pre-trained \texttt{TinyBERT\_General\_4L\_312D} model from the Hugging Face \texttt{transformers} library. The model is adapted to our specific task by adjusting the number of output labels to correspond to the categories in our dataset.

\subsubsection{Dataset preprocess}
In the context of our dataset, which primarily consists of about 5500 images of insects , a critical aspect of classification is identifying whether an insect belongs to a pest species. This determination hinges not only on the species of the insect but also significantly on its interaction with the image background, particularly in agricultural settings. LLAVA generates comprehensive text descriptions that encapsulate these crucial details based on image set.

When LLAVA analyzes each image, it goes beyond mere species recognition. It delves into the contextual relationship between the insect and its surroundings and the generated textual descriptions, therefore, contain nuanced information that describes not just the insect but also its activities and surroundings. For example, a description might include details such as "...a bee standing on a purple flower..." or "...a blue blanket with a small hole in it, possibly caused by a bug..." providing critical insights into the potential impact of the insect on the environment.
    \begin{figure}[h]
        \centering
        \includegraphics[width=1\linewidth]{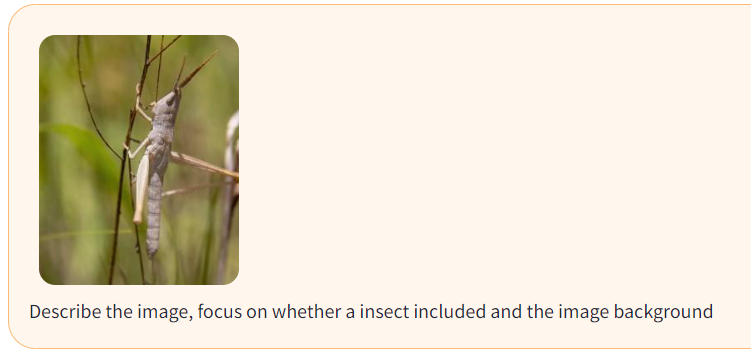}
        \caption{LLAVA prompt}
        \label{fig:llava}
    \end{figure}

This enriched data, once preprocessed and tokenized using TinyBERT's tokenizer, offers a detailed and contextual basis for training our model. By incorporating both the identification of the insect species and the contextual understanding of its role in the environment, our model is better equipped to differentiate pests from non-pests, leading to more effective and practical applications in agricultural and ecological settings.

\subsubsection{Training Procedure}
The fine-tuning of TinyBERT is carried out over a small number of epochs, considering the model's efficiency and our dataset's size. We employ the AdamW optimizer with a learning rate of \(5 \times 10^{-5}\) and cross-entropy loss to guide the training process. The model is trained using a custom DataLoader, ensuring efficient batch processing and shuffling of the training data. Below is a loss/accurate rate vs Epoch for a 10 epochs training for Finetuned TinyBert:
    \begin{figure}[H]
        \centering
        \includegraphics[width=1\linewidth]{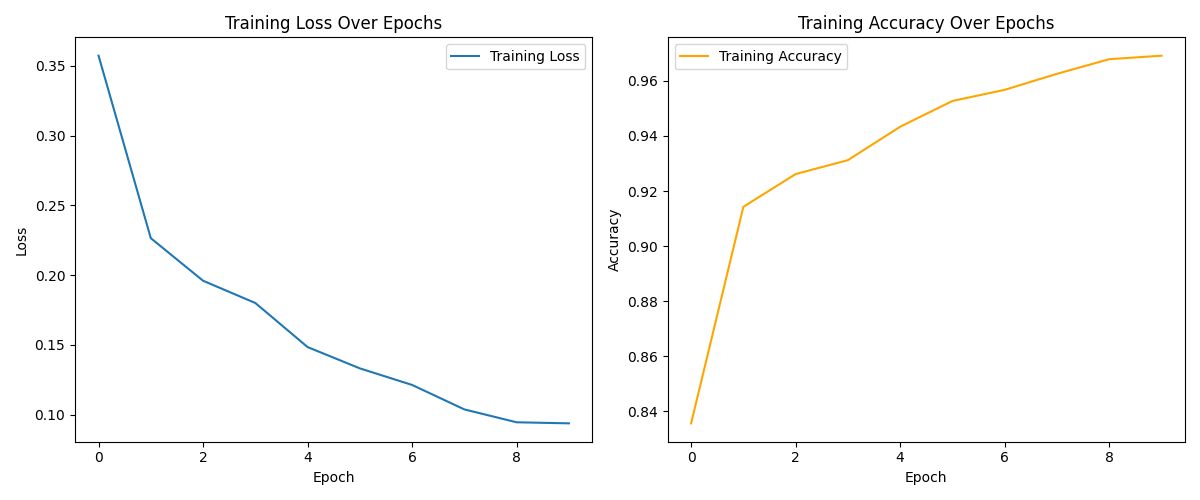}
        \caption{Loss/Accurate-Epoch}
        \label{fig:loss}
    \end{figure}
\subsubsection{Performance Metrics}
Model performance is evaluated using accuracy, a crucial metric given our task's classification nature. The accuracy is calculated on a separate test set, providing an objective measure of the model's ability to generalize to new data. This metric, along with qualitative analysis of model predictions, guides our assessment of the fine-tuning process's effectiveness.

In summary, the integration of TinyBERT allows our system to process and classify textual descriptions effectively, complementing the visual features extracted by the CNN-ResNet model. This dual approach, leveraging both image and text data, enhances the overall performance and applicability of our framework in multimodal classification tasks.

\subsection{Model Fusion Strategy}
Our ensemble learning framework is designed to leverage the distinct predictive strengths of different models, aiming to enhance overall accuracy. We integrate a CNN+ResNet model with a fine-tuned Tiny Bert for NLP tasks, applying a weighted average, linear regression, and random forest models for fusion.

\subsubsection{Weighted Average}
We calculate ensemble weights based on the accuracy metrics of the NLP and CV models to form a weighted average. The weights, \( w_{\text{NLP}} \) for the NLP model and \( w_{\text{CV}} \) for the CV model, are computed as follows:

\begin{align*}
w_{\text{NLP}} &= \frac{\text{Accuracy}_{\text{NLP}}}{\text{Accuracy}_{\text{NLP}} + \text{Accuracy}_{\text{CV}}} \\
w_{\text{CV}} &= \frac{\text{Accuracy}_{\text{CV}}}{\text{Accuracy}_{\text{NLP}} + \text{Accuracy}_{\text{CV}}}
\end{align*}

\subsubsection{Linear Regression Model}
A linear regression model is employed to process the outputs of the CNN+ResNet and Tiny Bert models. It is trained on the scores generated from these models and aims to predict the probability of the positive class:

\begin{itemize}
    \item The model is trained using scores as features and ground truth labels as the target.
    \item It predicts continuous scores on the test data, which are then thresholded at 0.5 to generate binary classifications.
    \item The accuracy of the binary predictions is assessed by comparison with the test labels.
\end{itemize}

\subsubsection{Random Forest Model}
We further employ a random forest classifier as part of our ensemble:

\begin{itemize}
    \item The classifier is trained on the same scores from the CNN+ResNet and Tiny Bert models' trainset.
    \item It predicts class probabilities for the test data.
    \item Probabilities are thresholded at 0.5 to create binary predictions.
    \item Model accuracy is determined by how well these predictions match the ground truth labels.
\end{itemize}

\section{Main Result}
Following the experimentation methods demonstrated above, as a result, we successfully integrated the linguistic and visual modalities to enhance pest detection: we processed textual data using the cutting-edge BERT model to capture contextual dependencies within pest description; concurrently, the ResNet model aided in the extraction of more complicated patterns from the visual input. And the merger of these elements was accomplished using multiple different concatenation methods that were specially tuned to capitalize on the particular strengths of each modality. In this section, we are going to take a glance at the final outcomes of our integrated model and analyze them using a visualized ROC \& AUC diagram.

\subsubsection{Performance Metrics}
In the field of pest identification, where the cost of false negatives can be significant (e.g., a single species of unidentified pest could leave thousands acres of field in threat from their erosion). Therefore, a predictive model's ability to distinguish between classes accurately is critical, and how to measure such ability is worth consideration. In our study, we used the Receiver Operating Characteristic (ROC) curve and the Area Under the Curve (AUC) to evaluate model performance.

The ROC curve is a graphical representation of a binary classifier system's diagnostic capacity by comparing the True Positive Rate (TPR) versus the False Positive Rate (FPR) at various threshold values. The TPR, also known as sensitivity or recall, calculates the percentage of true positives that are correctly identified as such. The FPR, on the other hand, is the percentage of true negatives that are misidentified as positives. A model with a TPR of 1 and an FPR of 0 is ideal, but in practice, a trade-off will always be made.

As for the AUC. by measuring the area under the ROC curve, it gives a single, aggregate measure of performance across all categorization criteria. AUC of 1 indicates a flawless model; AUC of 0.5 indicates no discriminative capacity, which is similar to random guessing. Following is our outcome, demonstrated in a Figure 4, including the ROC curve and AUC value of our two original models and several different model using various integrating approach.

    \begin{figure}[H]
        \centering
        \includegraphics[width=1\linewidth]{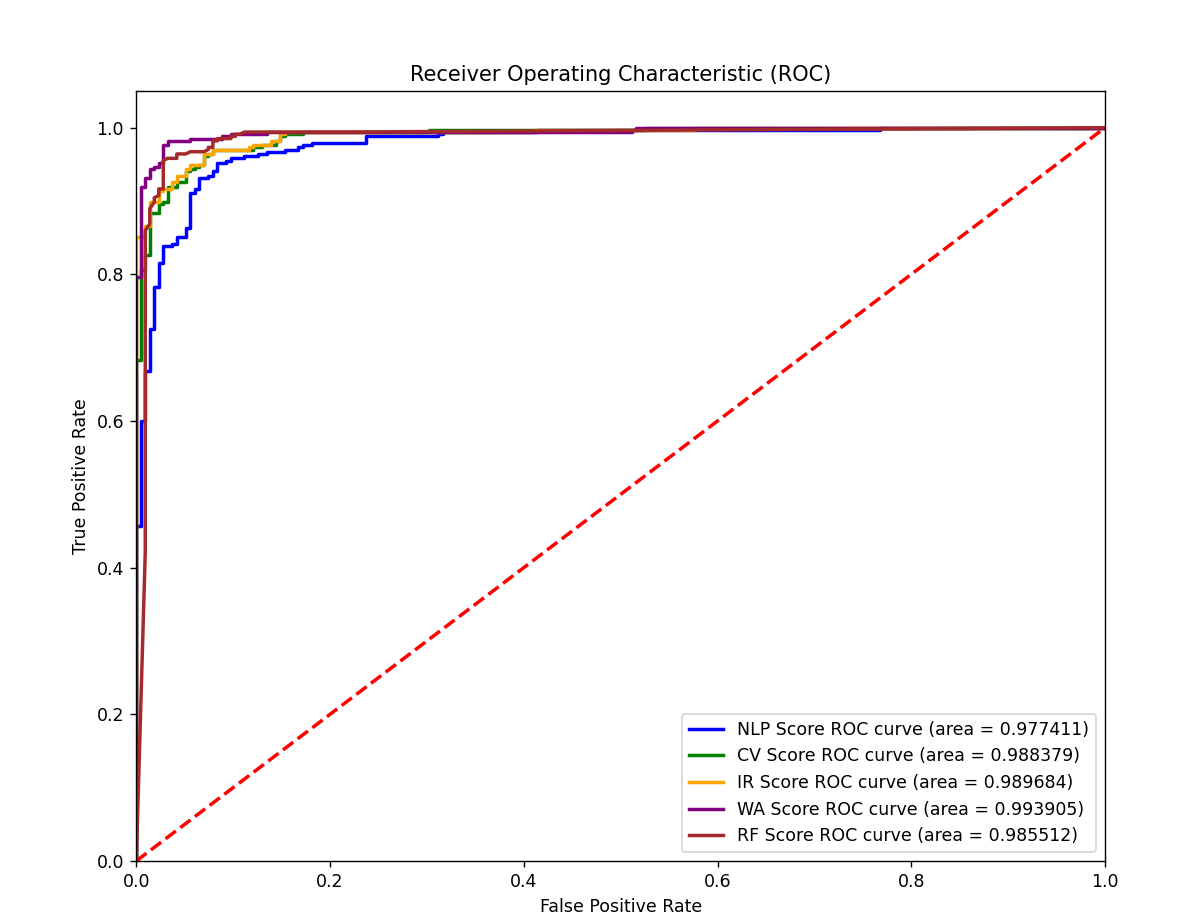}
        \caption{ROC and AUC for different models}
        \label{fig:Roc}
    \end{figure}

\subsubsection{Linear Regression (LR) Score Analysis}
The linear regression model produced a ROC curve with an AUC of 0.977 when trained using the combined feature set acquired from the NLP and CV scores. This displays a high level of discriminative ability, showing that the model can distinguish between the two classes reliably. The coefficients in the regression equation linked with the NLP and CV features highlighted the relative importance of each modality in predicting the outcome.
\subsubsection{Weighted Average (WA) Score Analysis}
The WA method provides a nuanced approach to prediction by combining accuracy-weighted ratings from both NLP and CV models. The estimated weights were directly proportionate to the NLP and CV models' individual accuracies. This approach generated a ROC curve with an impressive AUC of 0.994, indicating its superior performance in test set assessments.
\subsubsection{Random Forest (RF) Score Analysis}
We also used the RF model to , which operates by constructing multiple decision trees during training and outputting the class that is the majority vote of the individual trees \cite{BELGIU201624}. And it indeed had a significant discriminative capacity with an AUC of 0.985. The ROC curve linked with this model proved its robustness in accurately identifying the test data.

\subsubsection{Model Comparison}
A comparison found that, while all models performed admirably, the WA model had the greatest AUC, closely followed by the RF model. We were able to analyze the trade-offs between model complexity and performance using this comparison, with the WA model appearing as a promising method due to its balance of simplicity and high accuracy.

According to the comparative ROC analysis, the combination of NLP and CV modalities via a weighted average technique produced the greatest performance among all these tested methodologies. And such a fact implies that, while each modality has qualities of its own, their combination in various approaches can often lead to greater prediction capabilities. The constant AUC ratings above 0.97 for all models imply that each model can distinguish between the presence and absence of pests. The modest differences in AUC values among the models emphasize the impact of alternative feature representations and model topologies on the overall performance of the classifier.

\section{Conclusion}
According to the experiment outcomes illustrated above, conducted following the fine-tuning and fusion strategies, we now have confidence to conclude that using text-image multimodal models could enhance the performance of pest detection.

The improved performance of multimodal models that integrate textual and visual data can be attributed to synergistic augmentation of data representation, a phenomenon in which each modality contributes unique and complementary information that would otherwise be absent or overlooked if considered separately \cite{9498415}. On the one hand, textual descriptions elucidate behavioral patterns (such as granivorous or frugivorous tendencies), environmental contexts (such as the presence of leaves indicated by nibbled leaves), and temporal markers (denoting seasonal or diurnal cycles relevant to the observation), allowing for the differentiation of species with similar appearance and divergent ethological traits. 

Visual data, on the other hand, provides important morphological subtleties and conspicuous pattern recognition elements required for emphasizing fine-grained details. The intersection of various modalities results in a more comprehensive and nuanced data interpretation framework, which improves the model's capacity to distinguish and categorize with more accuracy and precision.

What's more, the convergence of textual and visual information serves not only as a means of broadening data representation, but also as a strategic approach to mitigating the impact of unimodal noise - extraneous or non-informative elements inherent in each individual modality. Each data type, whether text or image, is vulnerable to informational noise or extraneous aspects, which can obscure the underlying patterns required for accurate classification. By cross-referencing and correlating these disparate data streams, the multimodal paradigm intrinsically favors the intersection of consistent and prominent elements present in both modalities. This methodological synergy successfully filters out unimodal noise, improving signal-to-noise ratio. As a result, because it capitalizes on the inherent strengths of each data source while limiting their unique limitations, this integrated approach develops a more robust and precise prediction model.

\subsubsection{Future Improvement}
Our model's exceptional performance, as evidenced by an AUC close to 0.98, can be due in part to the deliberate use of dropout layers, a strategy used to reduce overfitting. Overfitting is a common problem in which a model becomes overly reliant on the training data, limiting its capacity to generalize. Dropout layers aid in model generalizability by randomly deactivating neurons during training. Furthermore, the dataset's small size and good quality are likely to have aided this level of performance. While a smaller, cleaner dataset helps to simplify the learning process and achieve unambiguous class distinctions, it's vital to note that it may not fully represent the model's performance in more complicated, real-world circumstances.

And in the foreseeable future, there are quite a lot of promising aspects for improvements. For example, we should increase the dataset size, which is critical for improving the model's accuracy and generalizability. A larger dataset with a broader range of pest species, environmental circumstances, and geographic locations can give the model with a deeper and more thorough training ground. This growth should not just focus on quantity but also on data diversity, including unusual and atypical insect infestation situations. A more diversified dataset would let the model to learn more complicated patterns while reducing the risk of overfitting, enhancing its performance in real-world circumstances.

Besides, using advanced data augmentation techniques can improve the model's robustness greatly. The reason why we argue so is that Advanced techniques for picture data, including as geometric modifications, photometric augmentations, and generative adversarial networks (GANs), can generate a variety of tough scenarios for the model to learn from \cite{mariani2018bagan}. And natural language processing techniques such as paraphrase, synonym replacement, and even producing synthetic text descriptions can provide a richer and more varied linguistic dataset for textual data. Overall, these enriched datasets can help the model better understand the complexities of pest detection and make it more resilient to changes in real-world data.

And finally, we could utilize integrating cross-modal attention mechanisms. As far as we're concerned, such utlization can improve the model's capacity to focus on relevant features of both text and visual input and such methods allow the model to recognize and prioritize the most informative information from each modality in real time, resulting in more accurate and contextually relevant predictions. This method is especially useful in complex settings where some textual properties may strongly correspond with specific visual patterns, necessitating the use of a model that can intelligently explore and integrate these multimodal inputs.

After all, you could access our project repository via:
\textbf{github.com/shirou10086/Deeplearning-NYU-2023}

\bibliography{aaai23}

\end{document}